\documentclass[letterpaper, 10pt, conference]{ieeeconf}      
\pdfoutput=1

\IEEEoverridecommandlockouts                              

\usepackage{graphics} 
\usepackage{epsfig} 
\usepackage{mathptmx} 
\usepackage{times} 
\usepackage{amsmath} 
\usepackage{amssymb}  
\usepackage{url}
\usepackage{booktabs} 
\usepackage{float}
\usepackage{placeins}
\usepackage{booktabs}
\usepackage{siunitx}
\usepackage{multicol}
\newcommand{\squeezeup}{\vspace{-2.5mm}}
\usepackage[ruled, shortend, linesnumbered]{algorithm2e}

\usepackage{breqn}
\title{\LARGE \bf Speeding Up Iterative Closest Point Using Stochastic Gradient Descent}
\author{Fahira Afzal Maken$^{1,*}$, Fabio Ramos$^{1,2}$ and Lionel Ott$^1$
\thanks{$^*$ Corresponding author: fafz3958@uni.sydney.edu.au}
\thanks{$^1$ School of Computer Science, The University of Sydney, Australia}
\thanks{$^2$ NVIDIA, USA}%
}

\begin{document}

\maketitle
\thispagestyle{empty}
\pagestyle{empty}
	
\begin{abstract}
    Sensors producing 3D point clouds such as 3D laser scanners and RGB-D cameras are widely used in robotics, be it for autonomous driving or manipulation. Aligning point clouds produced by these sensors is a vital component in such applications to perform tasks such as model registration, pose estimation, and SLAM. Iterative closest point (ICP) is the most widely used method for this task, due to its simplicity and efficiency. In this paper we propose a novel method which solves the optimisation problem posed by ICP using stochastic gradient descent (SGD). Using SGD allows us to improve the convergence speed of ICP without sacrificing solution quality. Experiments using Kinect as well as Velodyne data show that, our proposed method is faster than existing methods, while obtaining solutions comparable to standard ICP. An additional benefit is robustness to parameters when processing data from different sensors.
\end{abstract}

\section{Introduction}

Current 3D laser scanners and RGB-D cameras produce large 3D point clouds at a high frequency. For many applications the first step is to register or align pairs of such point clouds with each other. Accurately estimating the relative transformation between two such point clouds is paramount in applications such as SLAM, where it serves to constrain the pose graph, or in manipulation to perform scene reconstruction for future grasp planning. A widely used method for this, thanks to its simplicity and efficiency, is iterative closest point (ICP)~\cite{besl_1992}.

ICP takes two point clouds, source and reference, as input with the goal of estimating the transformation that aligns the two point clouds. Using an initial transformation guess, ICP finds corresponding points in the two point clouds and updates the transformation in order to minimise the Euclidean distance between the correspondences. These steps are repeatedly executed until convergence is achieved. The simplicity of ICP stems from the fact that the transformation is estimated solely on the basis of distances between points in the two point clouds. This is in contrast to other approaches which rely on features extracted from the point clouds in order to perform the data association.

The quality of an ICP solution depends on the accuracy of the initial guess and the overlap between the two point sets. The initial guess is typically obtained from a motion model using data from an inertial measurement unit (IMU), odometer, etc. An inaccurate initial guess, due to drift and limited precision of the motion model, can result in large alignment errors resulting in incorrect transformation estimates being produced.

Standard ICP uses all available data points in every iteration when computing the transformation between the two point clouds. With sensors capable of producing point clouds with tens or hundreds of thousands of points, this approach is computationally expensive. In this paper, we propose a novel approach to solving the optimisation problem of ICP by exploiting the recent advances in stochastic gradient descent (SGD) based optimisation. Instead of computing gradient information using the entire point clouds, as done by traditional ICP methods, we use smaller partial point clouds or mini-batches which can be processed faster and in constant time. While the gradient information obtained in this manner is noisier when compared to those obtained over the full point clouds, the gradients converge to the same result in expectation \cite{bottou2018}. This results in a method which solves the original ICP problem more quickly, without sacrificing stability or quality of the solution.

The contributions of this paper are as follows:
\begin{itemize}
    \item  A novel way to improve the speed and data efficiency of ICP to handle large scale point clouds;
    \item  A robust and efficient ICP method which is less sensitive to parameters when processing different data sources.
\end{itemize}

An implementation of the method described in this paper is available online\footnote{\url{https://bitbucket.org/fafz/sgd_icp/}}.

\section{Related Work}

Selecting corresponding points in the source and reference point clouds is an important step in ICP, as evidenced by the various ICP variants that have been proposed in the literature to improve this part of ICP. Luck et al.~\cite{Luck2000} use simulated annealing to obtain a good initial starting point. Masuda and Yokoyo~\cite{Masuda1995} combine a least median square estimator and ICP using random samples of points in each iteration. While this method uses random sub-sampling, the goal is to remove outliers from the point cloud processed by ICP and does not perform incremental updates of the transformation estimate as our proposed method does.

Chen and Medioni~\cite{Chen1992} propose a robust version of ICP which minimises the distance between points and planes (point-to-plane ICP) instead of the traditional point-to-point distance. In point-to-plane ICP, matching points in the two point clouds are determined by intersecting a ray from the source point in direction of source point's normal with the reference point's surface. This method is more robust to local minima when compared to standard ICP~\cite{Salvi2007}. Segal et al.~\cite{segal2009} combine point-to-point and point-to-plane ICP into a single probabilistic framework called Generalised-ICP (GICP). GICP is a plane-to-plane algorithm that models the local planar structure in both source and reference point clouds instead of just the reference's, as is typically done with the point-to-plane ICP. Comprehensive summaries of different ICP variants and their performance and properties can be found in \cite{Rusinkiewicz2001} and \cite{Pomerleau2015}.

Once the corresponding point pairs are selected the next challenge is to use this information to find the optimal transformation. Several optimisation techniques are used to minimise ICP's cost function. Fitzgibbon~ \cite{Fitzgibbon2003} presents a method which uses a non-linear Levenberg-Marquardt method \cite{Press1992} which combines batch gradient descent and Gauss-Newton to optimise the cost function.
Levine \cite{Levine1995} uses simulated annealing to minimise the cost function.

Another avenue of research are methods that improve the ability of ICP to handle an unknown degree of overlap between the point clouds. This includes the use of a trimmed square cost function to estimate optimal transformation \cite{Chetverikov2002}, rejection of corresponding point pairs based on a threshold defined by the standard deviation of point pair distances \cite{Zinsser}, or the use of a semi differential invariant for correspondence estimation \cite{Pajdla1995}.

Aligning point clouds and estimating relative transforms between them can be achieved using other methods as well, including \cite{srivatsan2017} and \cite{ArunSrivatsan}. These methods use distribution based filtering approaches to model the uncertainties in the transformation parameters.

Our proposed method does not introduce a novel formulation of ICP, rather it provides a different method of solving the underlying optimisation problem, in a manner that is computationally more efficient and less reliant on parameter tuning.

\section{Stochastic Gradient Descent ICP}

In the following subsection, we introduce the optimisation problem underlying ICP before describing our proposed method, SGD-ICP, which solves the ICP problem using stochastic gradient descent.

\subsection{Standard ICP}

Standard ICP aligns a pair of point clouds in two steps. In the first step, the source cloud is transformed with a current transform and then the correspondence between pairs of points in the transformed source and the reference point clouds are established on the basis of Euclidean distance. In the second step, a transformation which minimises the point-to-point Euclidean distance between the corresponding points is calculated. The cost function or loss $\mathcal L$ optimised by ICP in the second step is a function of $\theta = \{x, y, z, \text{roll}, \text{pitch}, \text{yaw}\}$ representing the translation and rotation needed to align the two point clouds, i.e.:
\begin{equation}
    \operatorname*{argmin}_\theta  \mathcal L(\theta) = \frac{1}{N}\sum_i^N  ||T_\theta s_i - r_i||^2,
    \label{eq:icp-objective}
\end{equation}
where \textit{N} represents the total number of points in the source cloud, $s_i$ is a point in the source cloud, $r_i$ is the corresponding point in the reference cloud, and $T_{\theta} \in \mathbb R^{4 \times 4}$ is a transformation matrix parametrised by the parameter vector $\theta$. $T_\theta$ can be decomposed into $t_{\theta^{1:3}}$ and $R_{\theta^{4:6}}$, where $t_{\theta^{1:3}} \in \mathbb R^{3 \times 1}$ is a translation vector and $R_{\theta^{4:6}} \in \mathbb R^{3 \times 3}$ is a rotation matrix. Rewriting \eqref{eq:icp-objective} using the decomposition of $T_{\theta}$, we obtain :
\begin{equation}
    \operatorname*{argmin}_\theta  \mathcal L(\theta) = \frac{1}{N}\sum_i^N  ||(R_{\theta^{4:6}}s_i+t_{\theta^{1:3}}) - r_i||^2
    \label{eq:icp-objective2}
\end{equation}

\subsection{Stochastic Gradient Descent ICP}

The optimisation problem in \eqref{eq:icp-objective} is usually solved using batch optimisation methods in ICP. However, batch optimisation algorithms scale badly to problems where a large number of data points have to be processed in each step. This motivates the use of stochastic gradient descent (SGD) to solve the optimisation step in ICP. Stochastic gradient descent computes a gradient based on a single point or small number of points, mini-batch, as opposed to the full point cloud, thus resulting in a computational cost that is independent of the size of the point cloud. While computing a gradient in such a manner results in a noisy estimate of the true gradient, one can prove convergence and optimality of SGD \cite{bottou2018}.

We can minimise the loss $\mathcal L(\theta)$ in \eqref{eq:icp-objective} using mini-batches containing $m$ points with SGD with the following update equation for the transformation parameters $\theta$:
\begin{equation}
    \theta_t = \theta_{t-1} - \alpha A \frac{1}{m}\sum_i^m \frac{\partial}{\partial \theta} \mathcal L(\theta),
    \label{eq:sgd-fomulation}
\end{equation}
where $\frac{\partial}{\partial \theta}\mathcal L(\theta)$ are the partial derivatives of the loss with respect to the individual parameters. $\theta_t$ and $\theta_{t-1}$ are the values of the transformation parameters at the current and previous iteration respectively. The matrix $A$ acts as a pre-conditioner and is assumed to be the identity matrix in our case and the learning rate or step-size $\alpha$ dictates how quickly parameter values change. The step-size can be set according to various schedules, including fixed step size and adaptive schemes such as ADAM~\cite{kingma2015}. 

The full update rule for the parameter vector $\theta$ has the following form:
\begin{equation}
    \theta_{t} = \theta_{t-1} - \alpha \frac{1}{m} \sum_i^m(\frac{\partial}{\partial {\theta^{1:6}}} \left\Vert (R_{\theta^{4:6}}s_i+t_{\theta^{1:3}}) - r_i \right\Vert^2)
    \label{eq:sgd-icp-cost}
\end{equation}\par
For translational update, $\frac{\partial}{\partial \theta^{1:3}} \mathcal L(\theta)$ is independent of the rotational components, thus $\frac{\partial R_{\theta{4:6}}}{\partial \theta^{1:3}} = 0$. Similarly, for rotation angle updates, $\frac{\partial}{\partial \theta^{4:6}} \mathcal L(\theta)$ does not depend on translational parameters, resulting in $\frac{\partial t_{\theta{1:3}}}{\partial\theta^{4:6}} =0$. Equation \eqref{eq:sgd-icp-cost} can further be made more explicit by splitting it into translational updates, $\theta^{1:3} = \{x, y, z\}$
\begin{equation}
    \theta^{1:3}_t = \theta^{1:3}_{t-1} - \alpha \frac{1}{2m} \sum_i^m ((R_{\theta^{4:6}}s_i+t_{\theta^{1:3}}) - r_i ) \frac{\partial t_{\theta^{1:3}}}{\partial\theta^{1:3}},
    \label{eq:sgd-icp-update-translation}
\end{equation}
and rotation angle updates $\theta^{4:6} = \{\text{roll}, \text{pitch}, \text{yaw}\}$
\begin{equation}
    \theta^{4:6}_t = \theta^{4:6}_{t-1} - \alpha \frac{1}{2m} \sum_i^m (( R_{\theta^{4:6}}s_i+t_{\theta^{1:3}}) - r_i) \frac{\partial R_{\theta^{4:6}}}{\partial \theta^{4:6}} s_i,
    \label{eq:sgd-icp-update-rotation}
\end{equation}
where $\frac{\partial t_{\theta{1:3}}}{\partial\theta^{1:3}}$ are $\mathbb R^{3 \times 1}$ vectors and $\frac{\partial R_{\theta{4:6}}}{\partial \theta^{4:6}}$ are $\mathbb R^{3 \times 3}$ matrices. For example, $\frac{\partial t_{\theta{1:3}}}{\partial x}$, $\frac{\partial t_{\theta{1:3}}}{\partial y}$, and $\frac{\partial t_{\theta{1:3}}}{\partial z}$ are
$
\begin{bmatrix}
1 & 0 & 0
\end{bmatrix}^\intercal 
$,
$
\begin{bmatrix}
0 &1 & 0
\end{bmatrix}^\intercal
$, and 
$
\begin{bmatrix}
0 & 0 & 1
\end{bmatrix}^\intercal 
$
 respectively and $\frac{\partial R_{\theta{4:6}}}{\partial roll}$ is
$
\begin{bmatrix}
    $$\frac{\partial r_{11}}{\partial roll}$$ & $$\frac{\partial r_{12}}{\partial roll}$$ & $$\frac{\partial r_{13}}{\partial roll}$$\\
    $$\frac{\partial r_{21}}{\partial roll}$$ & $$\frac{\partial r_{22}}{\partial roll}$$ & $$\frac{\partial r_{23}}{\partial roll}$$\\
     $$\frac{\partial r_{31}}{\partial roll}$$ &  $$\frac{\partial r_{32}}{\partial roll}$$ &  $$\frac{\partial r_{33}}{\partial roll}$$
\end{bmatrix}
$. $\frac{\partial R_{\theta{4:6}}}{\partial pitch}$ and $\frac{\partial R_{\theta{4:6}}}{\partial yaw}$ can be derived in a similar manner. The individual components of the rotation matrix ($r_{11}$ to $r_{33}$) can be found in ~\cite{Chen1992}.\par
In SGD-ICP, a mini-batch is obtained by picking a random set of $m$ points from the source point cloud. Each point in the mini-batch is then transformed with the current transformation matrix, i.e., $s_i\textprime = T_{\theta_{t-1}}s_i$. The corresponding points to these transformed points $(s_i\textprime)$ in the reference point cloud ($R_{ef}$) are found using the point-to-point distance metric:

\begin{equation}
    \text{point-to-point}(s_i\textprime, R_{ef}) = \min_{r \in R_{ef}} \left\Vert r - s_i\textprime \right\Vert
\end{equation}


In case of partial overlap between the two point clouds, we use a maximum correspondence threshold $d_{\text{max}}$ to reject unlikely matches. Using the point pairs obtained in this manner, we iteratively minimise the Euclidean distance between the two clouds using \eqref{eq:sgd-icp-cost}.

Algorithm \ref{alg:sgd-icp} shows the steps performed by our method. First a new mini-batch is obtained from the source cloud in line 2. It is then transformed with the latest transformation matrix in line 3, then in lines 4 to 8 the corresponding points to those points in the reference cloud are found and stored. Once the corresponding pairs have been obtained, the stochastic gradient is computed and used to update the translation and rotation parameters of the transformation in lines 9 and 10. Finally, upon convergence the final set of parameters is returned in line 13.

In order to avoid bias in the mini-batch selection, sampling is performed without replacement until all points have been selected, at which point all points are added back to the sampling pool where they are re-considered for sampling. As the spatial extents of the point clouds can vary a lot, each point's position is scaled to $[0, 1]$. This results in the step-size not being dependant on the spatial extent of the point clouds being aligned. This scaling is undone before returning the final transformation.

Although this paper focuses on the point-point cost variant of ICP, it is in principle applicable for any other cost variant of ICP.


\begin{algorithm}[bt]
	\caption{SGD-ICP}
	\label{alg:sgd-icp}
	
	\SetAlgoLined
  
    \SetKwInOut{Input}{Input}
    \SetKwInOut{Output}{Output}     
    
    \Input{
        Two point clouds, source $S = \{s_i\}$ and reference $R_{ef} = \{ r_i \}$\\
        Initial transformation parameters: $\theta_0$\\
        Mini-batch size: $m$
    }
    \Output{Optimal transformation parameters: $\theta_t$}
    
	\While{not converged t $\leftarrow$ 1} 
	{
        Batch $\leftarrow $ pick a mini-batch cloud of size $m$ from $S$\\
        Batch\textprime $\leftarrow$ transform mini-batch with $\theta_{t-1}$\\
        Pairs $ \leftarrow \emptyset$\\
        \For{$s_i^\prime \in$ Batch\textprime}
        {
            $r_i \leftarrow$ closest point in $R_{ef}$ to $s_i^\prime$\\
            Pairs $\leftarrow$ Pairs $\cup \quad \{s_i^\prime, r_i\}$\\
        }
            
        $\theta^{1:3}_t \leftarrow \theta^{1:3}_{t-1} - \alpha \frac{1}{2m} \sum_{s_i^\prime, r_i \in \text{Pairs}} ( s_i^\prime - r_i) \frac{\partial(s_i^\prime - r_i)}{\partial \theta^{1:3}}$\\
        $\theta^{4:6}_t \leftarrow \theta^{4:6}_{t-1} - \alpha \frac{1}{2m} \sum_{s_i^\prime, r_i \in \text{Pairs}} (s_i^\prime - r_i) \frac{\partial(s_i^\prime - r_i)}{\partial \theta^{4:6}}$\\
        t $\leftarrow$ t+1\\
    }
    \Return{$\theta_t$}
\end{algorithm}
\vspace{-1em}

\section{Experiments}

The following experiments investigate the performance of SGD-ICP in terms of data-efficiency, run time and accuracy, robustness, as well as parameter sensitivity. We compare our proposed method against standard ICP and Generalized ICP (GICP)~\cite{segal2009}, using the PCL~\cite{rusu2011} implementations, as well as a point-to-point (LIB-POINT) and point-to-plane (LIB-PLANE) based ICP variants implemented in libpointmatcher as described in \cite{Pomerleau2011}. The libpointmatcher methods use random sub-sampling in order to improve the run-time of the methods at the expense of some solution quality.
We also investigate the effect the mini-batch size has on the performance of SGD-ICP in terms of accuracy and run-time.
The quality of the transformation estimates is measured separately for translation and rotation. Translational error is measured as the sum of Euclidean distance errors while rotational error is measured as the sum of the absolute angular differences.

Experiments are performed using the ETHZ-ASL Kinect Dataset\footnote{https://projects.asl.ethz.ch/datasets/doku.php} \cite{Pomerleau2011} which contains Kinect scans from an indoor scene with varying amount of clutter. The ground truth for this dataset is obtained by aligning the point clouds using standard ICP using the transform obtained from the Vicon system as the initial guess. As we are interested in pairwise transformations rather than tracking, this approach provides a better ground truth. The second dataset we use is the KITTI datset \cite{geiger2013} from which we use the Velodyne data. As no high accuracy ground truth is available for scan-to-scan alignment, we apply random transformations of up to a maximum translation motion of \SI{30}{\meter} and rotational motion of \SI{30}{\deg} onto Velodyne scans. 

In all the experiments, we use a value of $d_{\text{max}} = 0.5$ to discard incorrect correspondences, a mini-batch size of $m = 160$ points and a step size of $\alpha = 2$. These values work well in practice and no attempt was made to find the best performing ones. For libpointmatcher based methods, we use a sub-sampling ratio of $0.5$ to achieve good accuracy and speed. The experiments are performed on a desktop PC with an Intel Core i7-7700 CPU and 16 GB RAM. The source code implementation is single-threaded and does not employ any GPU acceleration.

\subsection{Data-Efficiency}

\begin{figure}[bt]
    \centering
    
	\includegraphics[width=\linewidth]{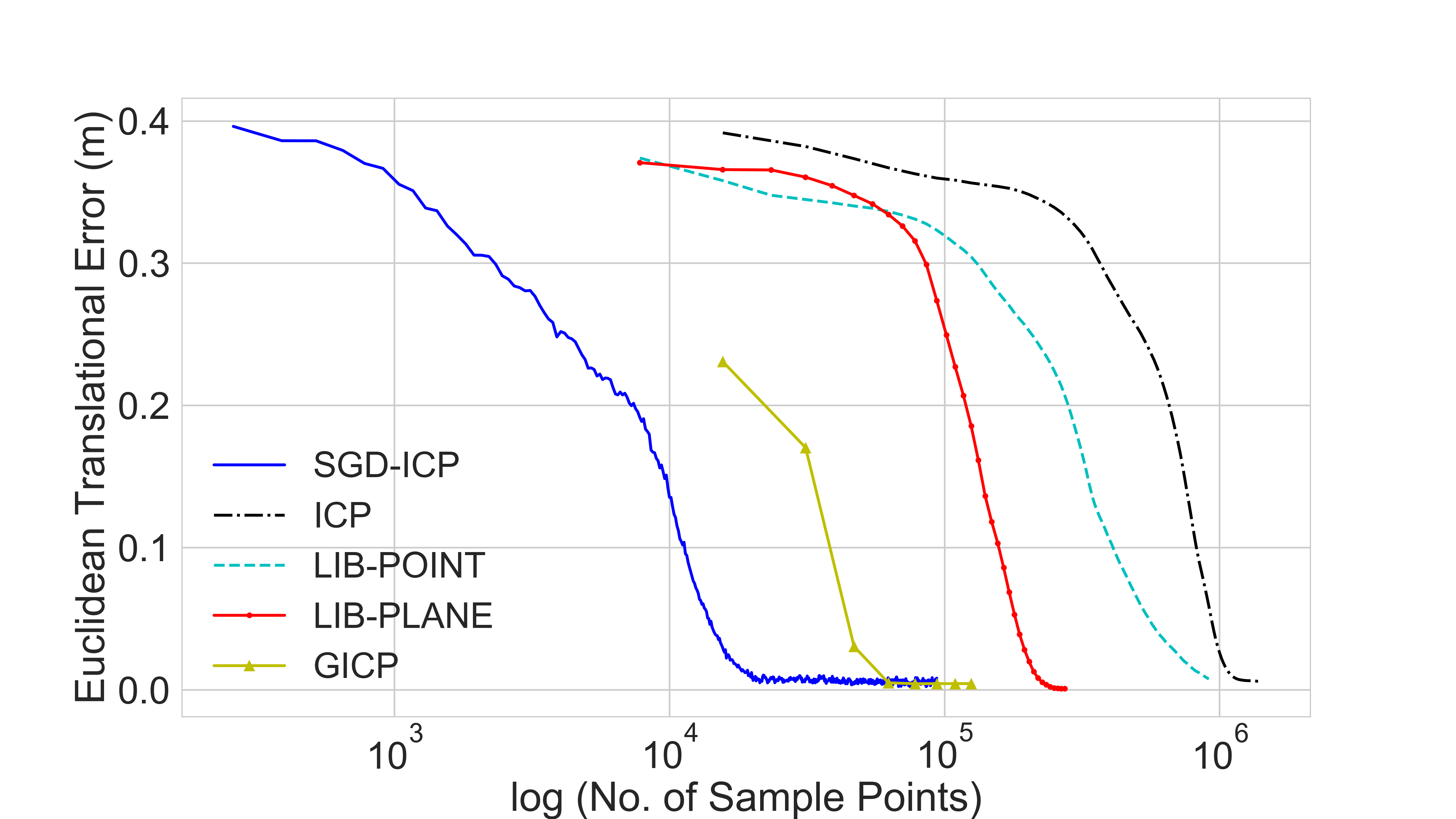}
	\caption{Comparison of the amount of points needed by different methods to achieve the same level of performance. The amount of points processed is shown in log-scale on the X-axis and clearly shows how SGD-ICP achieves the same result as other methods in roughly a single pass through the point cloud which contains 15646 points.}
	\label{fig:data-efficiency}
\end{figure}

We begin by evaluating how many data points the different methods need to process, to obtain a given level of solution quality. This is done by recording the number of points used by each method in each iteration until convergence is achieved. Figure~\ref{fig:data-efficiency} shows the number of points processed in log-scale along the X-axis with the translational error (in meters) along the Y-axis. The starting point of each line indicates the translational error and number of points used by each algorithm after one iteration.
This plot clearly shows how our method (SGD-ICP) requires significantly less points, equivalent to a single pass through the point cloud of 15646 points, than the other methods to converge. GICP needs less passes over the cloud than the other methods, which is explained by the more informative error used and a robust point correspondence method. However, as we will see this comes at the cost of run time. The other three methods all use a large amount of points, yet all five methods achieve comparable errors. From Figure~\ref{fig:data-efficiency}, we can additionally see that the libpointmatcher variants are more data efficient than standard ICP. This stems from the pre-processing done by these methods to the initial point clouds which includes randomly sub-sampling the original point clouds to reduce the total amount of points processed in each iteration.

This example demonstrates that the stochastic gradient estimates of our method are sufficient to achieve results comparable to other methods but at a fraction of the data being processed. In the next section we are going to see how this reduction in data usage of SGD-ICP translates to run-time efficiency and impact on solution accuracy.

\subsection{Solution Quality and Run-time}

To evaluate the quality of the solution and run-time of the algorithms, we use 1000 cloud pairs randomly selected from the medium-medium-fly ETHZ-ASL Kinect dataset. The average alignment error and run-time along with their standard deviations are shown in Table~\ref{tab:quality-run-time}. The translational and rotational error is presented in meters (m) and radians (rad) respectively with run-time in seconds (sec).

\begin{table}[bt]
	\begin{center}
		\caption{Accuracy and Time Comparison of SGD-ICP against other Methods Using Kinect Dataset}
		\label{tab:quality-run-time}
		
		\begin{tabular}{lrrr}
		    \toprule
			Method & Translational & Rotational & Run-Time\\
			       & Error (m)   & Error (rad)& (sec)\\
            \midrule
			SGD-ICP   & $0.134 \pm 0.190$ & $0.123 \pm 0.143$ & $ 0.330 \pm  0.522$ \\
			ICP       & $0.104 \pm 0.162$ & $0.102 \pm 0.132$ & $ 5.156 \pm  2.325$ \\
			LIB-POINT & $0.279 \pm 0.390$ & $0.365 \pm 0.705$ & $ 6.273 \pm  3.050$ \\
			LIB-PLANE & $0.266 \pm 0.407$ & $0.362 \pm 0.755$ & $ 2.642 \pm  1.398$ \\
			GICP      & $0.069 \pm 0.164$ & $0.051 \pm 0.116$ & $84.065 \pm 44.438$ \\
			\bottomrule 
		\end{tabular}
	\end{center}
\end{table}

Looking at the errors we can see that, SGD-ICP obtains results equivalent to those of ICP. This is expected, as the optimisation problem solved by the two problems is identical. GICP obtains the solutions with the lowest error which can be attributed to the more informative error metric as well as point correspondence method. Finally, both libpointmatcher based methods (LIB-POINT and LIB-PLANE) obtain results with a larger error than standard ICP, which is part of the trade-off incurred by the pre-processing of these methods.\par 
Turning our attention to the run-time values, it is clear that SGD-ICP performs the best, being several times faster than standard ICP, while achieving the same error. This shows that, the reduction in number of points needed by SGD-ICP to converge translates directly into run-time efficiency. This also reveals that the run-time cost incurred to obtain the high quality results of GICP, being 40 times slower than standard ICP. A possible reason for this increase in run-time is the fact that, the computation of normal information can be costly, especially when done in combination with a full batch optimiser such as Broyden-Fletcher-Goldfarb-Shanno (BFGS)~\cite{AvrielMordecai2003} as used by GICP. Looking at the libpointmatcher methods, one can see that the reduction in quality has a beneficial impact on the run-time. 

While both our proposed method SGD-ICP and the LIB-POINT and LIB-PLANE methods process smaller parts of the original point clouds in each iteration, we can see the impact the fixed choice of points in LIB-POINT and LIB-PLANE has on the error. The fixed choice of points results in the points that might be informative being discarded, in contrast to our method, which has access to different parts of the point cloud in each iteration. This results in fast per-iteration time without losing possible information.

Another interesting point is the comparison of convergence speed of LIB-PLANE to standard ICP and LIB-POINT, both of which use a point-to-point based error metric. The faster convergence of LIB-PLANE indicates that plane-to-plane based ICP provides a substantial benefit over point-to-point based metrics. As SGD-ICP only addresses the way in which the optimisation is performed, it will be interesting to see how using the plane-to-plane metric performs with it.

\subsection{Impact of Error in Initial Transformation on Accuracy}

\begin{figure*}[bt]
    \centering

	\includegraphics[width=0.48\textwidth]{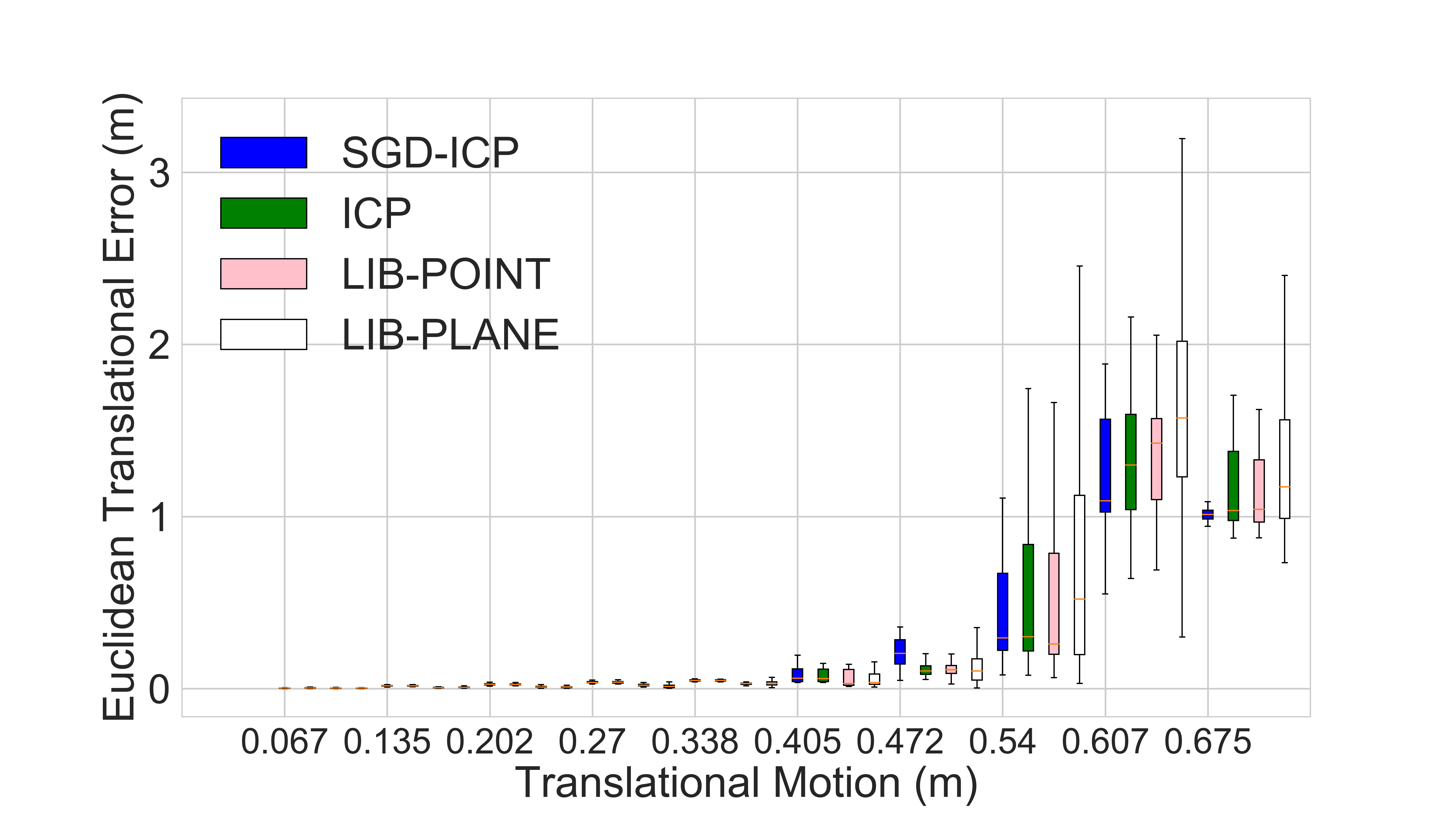}
	\hfill
  	\includegraphics[width=0.48\textwidth]{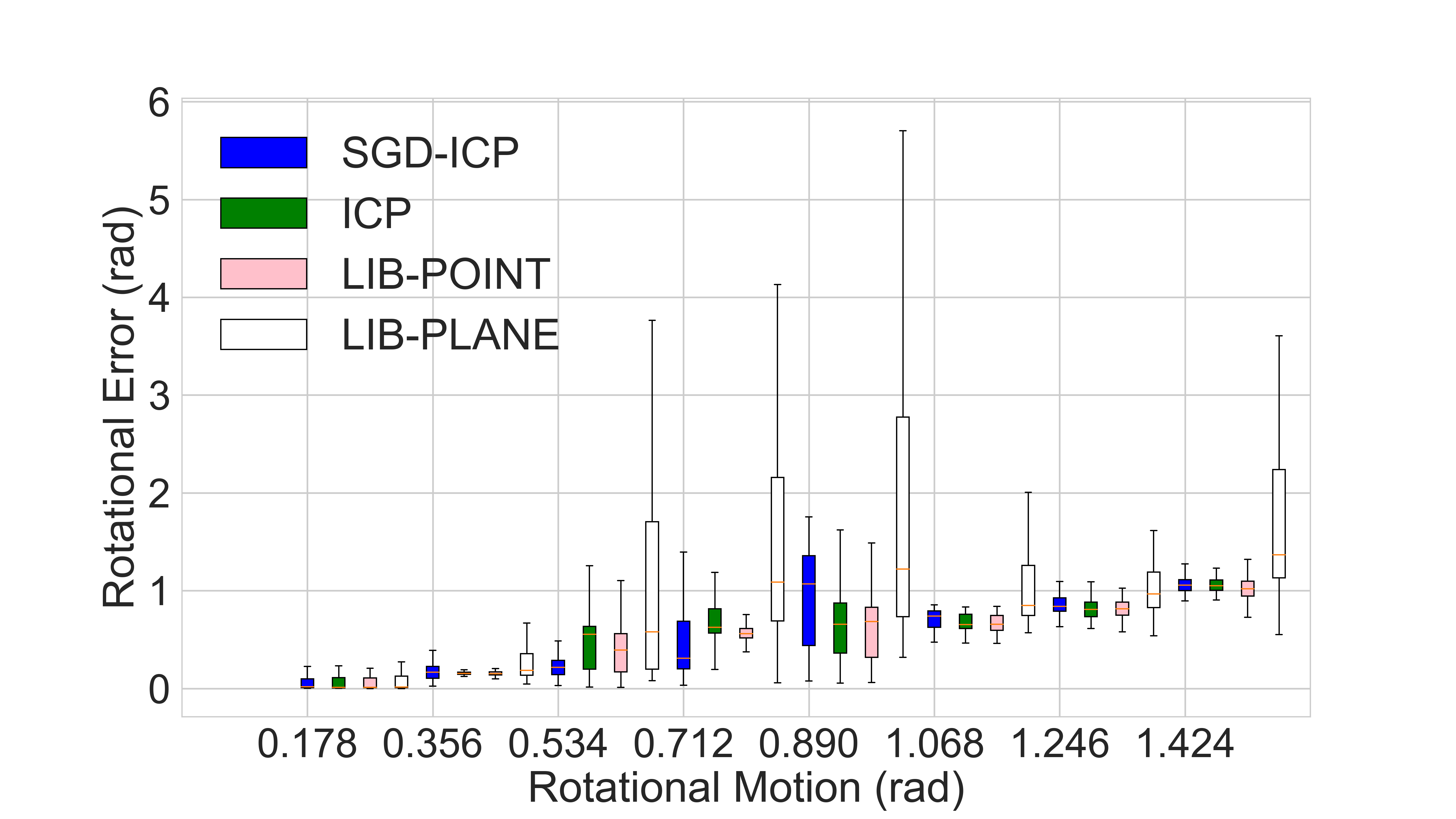}
	
	\caption{Comparison of the ability to handle errors in the initial transform by the different methods. The two plots show sensitivity to translational error (left) and rotational error (right). The X-axis shows the amount of error in the initial transform in meters and radians respectively.}
	\label{fig:initial-guess}
\end{figure*}

As our proposed method relies on stochastic gradients, this experiment aims to explore whether there is any instability arising from this. To this end, we analyse how the algorithms perform with various levels of mis-specification of the initial guess of the translation and rotation. We use the medium-medium-translation and medium-medium-rotation recordings of the ETHZ-ASL Kinect dataset. The possible pairings between point clouds are grouped into bins, based on the amount of translation or rotation needed to align them correctly. The maximum translation is \SI{0.675}{\meter} and the maximum rotation is \SI{1.75}{\radian}. These ranges are then split into ten evenly sized bins and 100 point cloud pairs for each of the ten bins are selected. For rotations the last two bins are excluded, as very few examples exist for these.

\begin{figure}[bt]
    \centering
    \includegraphics[width=0.9\linewidth]{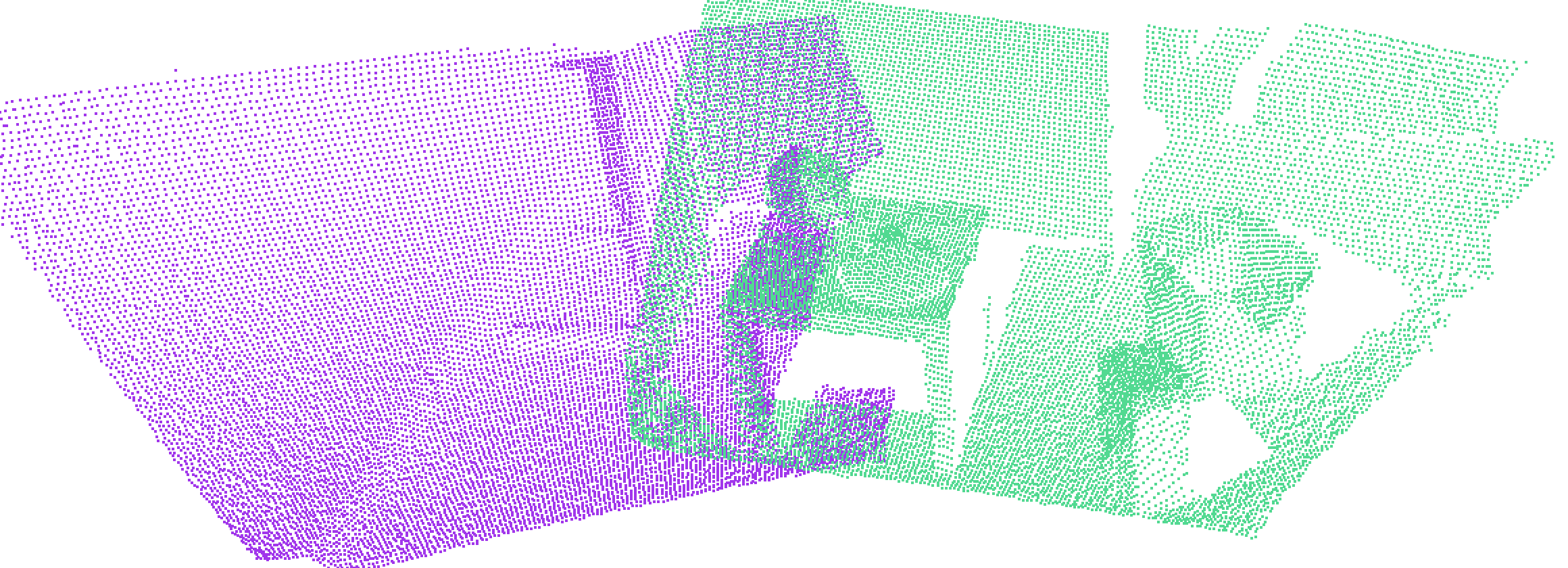}
    \includegraphics[width=0.9\linewidth]{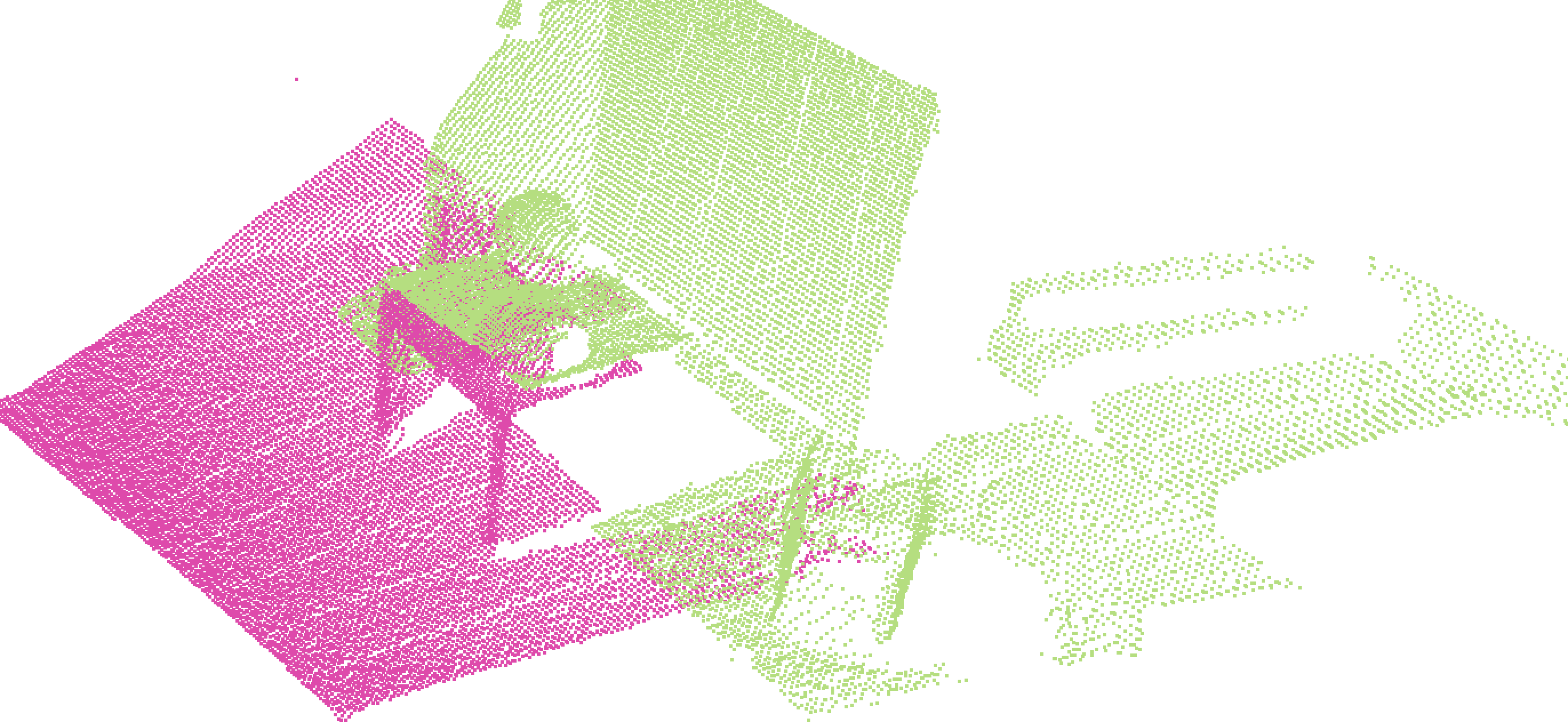}
    \caption{Example cloud pairs showing failure cases due to partial overlap resulting in under-constrained features between source (left) and reference (right) clouds.}
    \label{fig:partial_featureless_overlap}
\end{figure}

Figure~\ref{fig:initial-guess} shows the distribution of errors in translation (left) and rotation (right) as box plots. Along the X-axis the different error bins are shown while the Y-axis shows the error. For errors in translation, we can see that all methods perform without fault until roughly \SI{0.54}{\meter} error in translation, at which point all methods start to degrade in a similar manner. The reason for this is that, at that point the overlap between the scans becomes very small, resulting in many ambiguous solutions. For example single planar surface with no or few features hindering 
in finding the correct alignment, as shown in the two examples representing the ground truth alignment of a reference cloud on the right and source cloud on the left in Figure \ref{fig:partial_featureless_overlap}.

The comparison of the impact of error in the initial rotation (right hand figure \ref{fig:initial-guess}) shows the amount of initial rotational offset along the X-axis. The behaviour here is similar to the translational case in that, all methods start to degrade at a similar point, roughly at \SI{1.0} {\radian}. Again, similar to the previous test this coincides with a low overlap between the clouds. Equivalent performance of SGD-ICP to that of other methods shows that the stochastic nature of SGD-ICP does not result in a reduced ability to handle initial mis-specification of transformations.
\setlength{\textfloatsep}{1\baselineskip plus 0.2\baselineskip minus 0.2\baselineskip}
\squeezeup
\subsection{Effect of Batch size on the Solution Quality and Run-time}
 To investigate the effect of a batch-size on the solution quality and run-time, we use the same dataset as is used in section IV B. 
 This experiment is performed with a fixed step-size (vanilla-SGD) $\alpha = 2$ as well as an adaptive step-size ADAM~\cite{kingma2015} (Adam-SGD).
 
\begin{figure*}

	\includegraphics[width=0.46\textwidth]{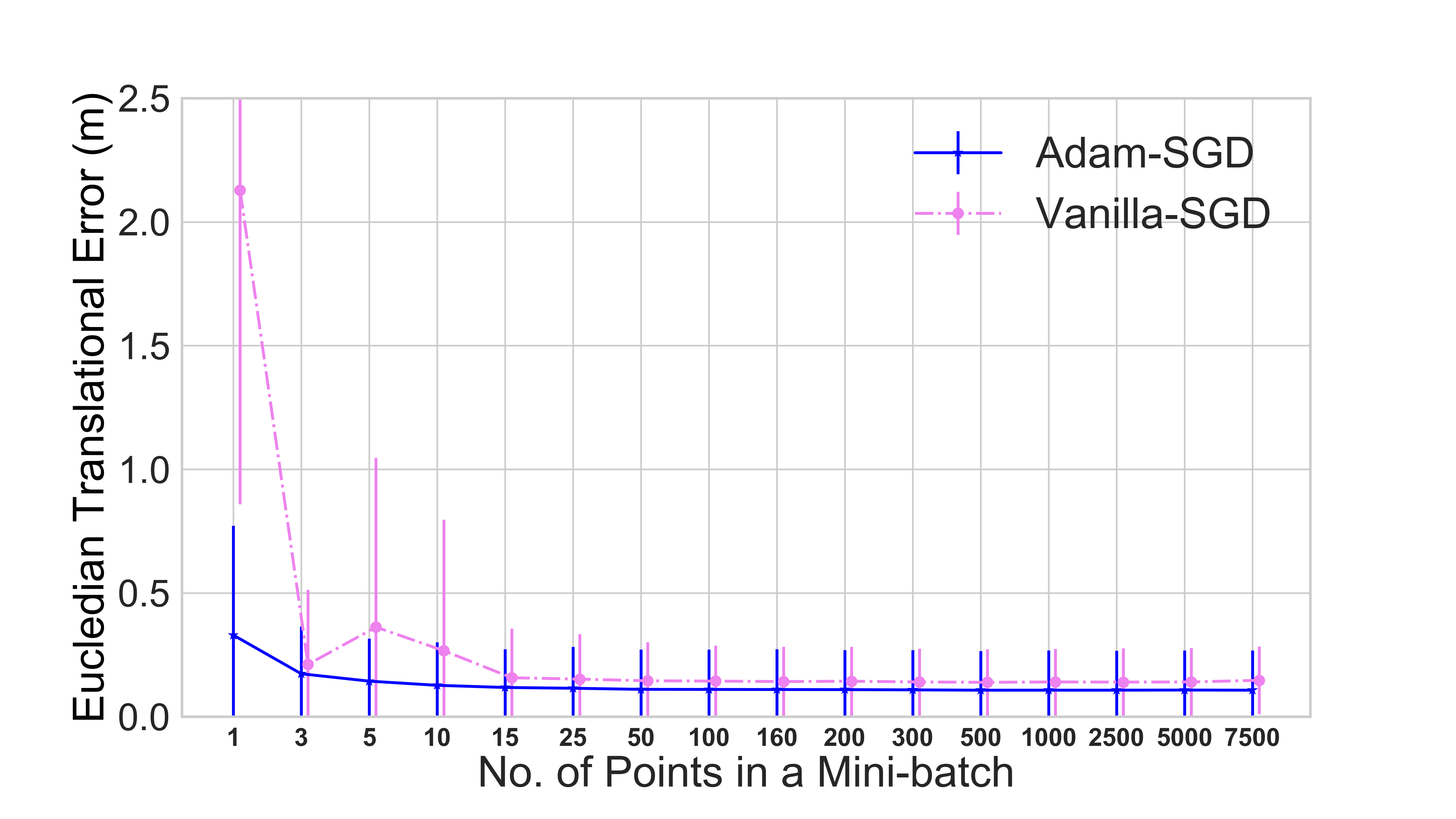}
	\hfill
  	\includegraphics[width=0.46\textwidth]{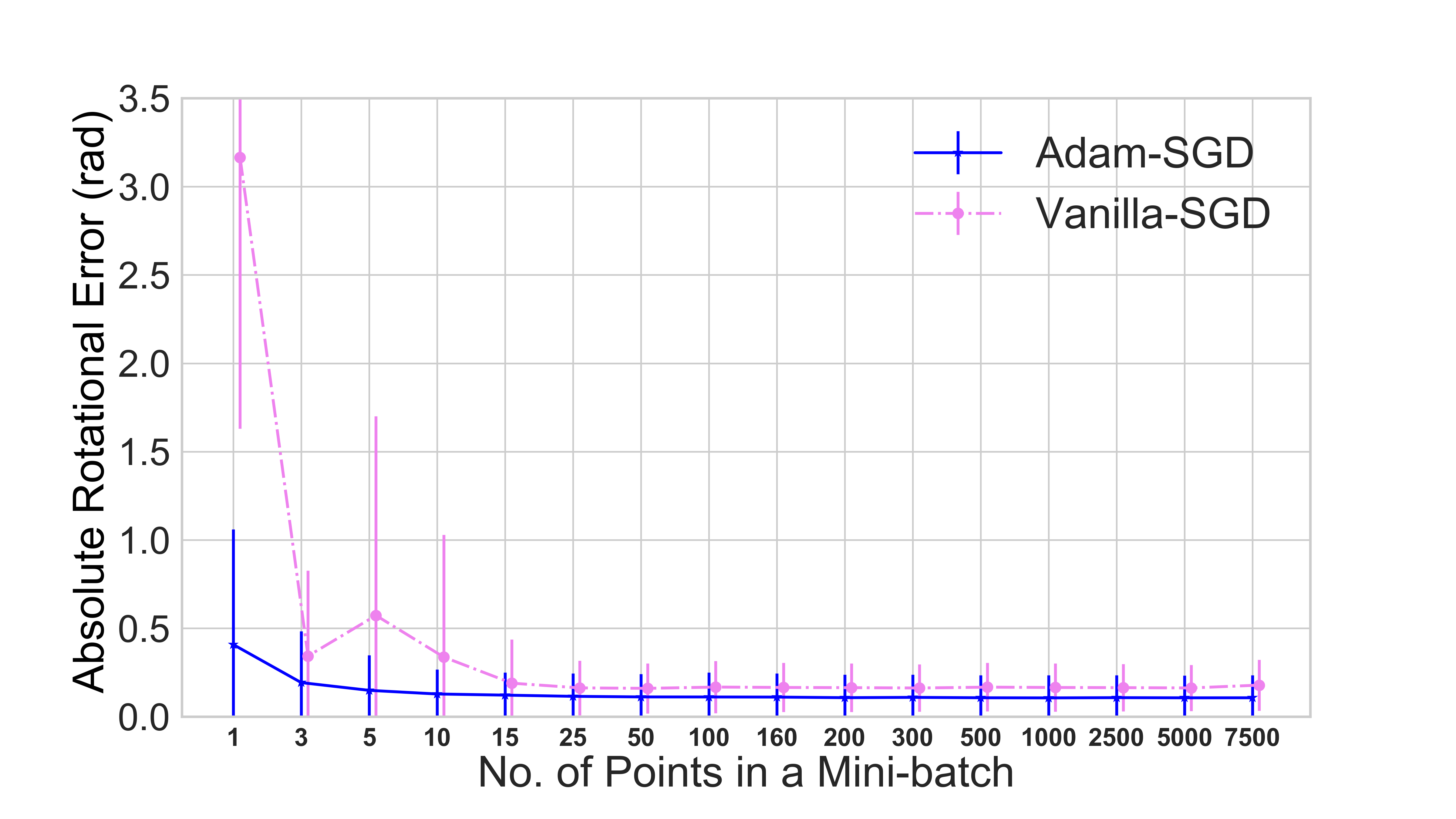}
  	\includegraphics[width=0.46\textwidth]{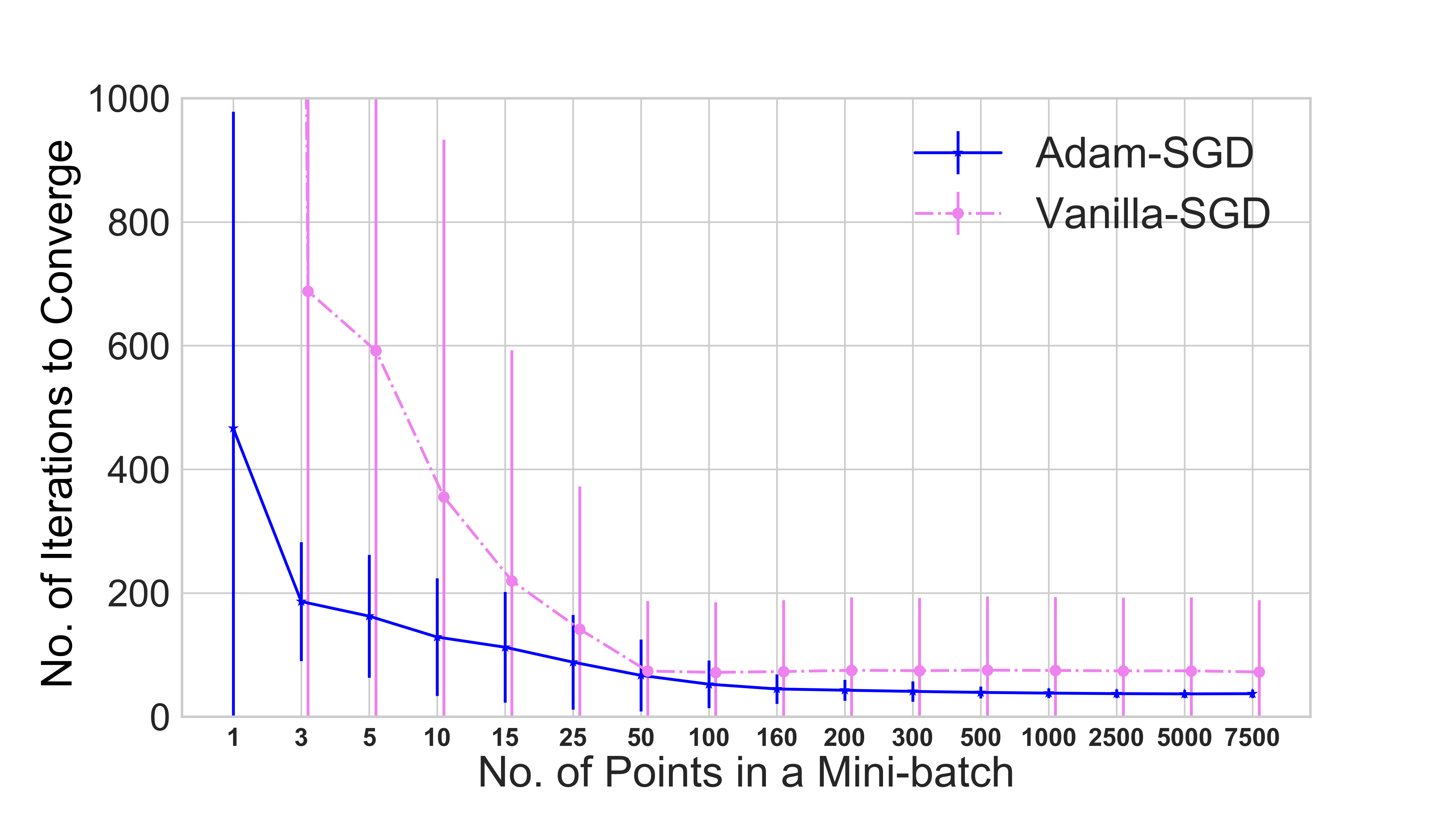}
	\hfill
  	\includegraphics[width=0.46\textwidth]{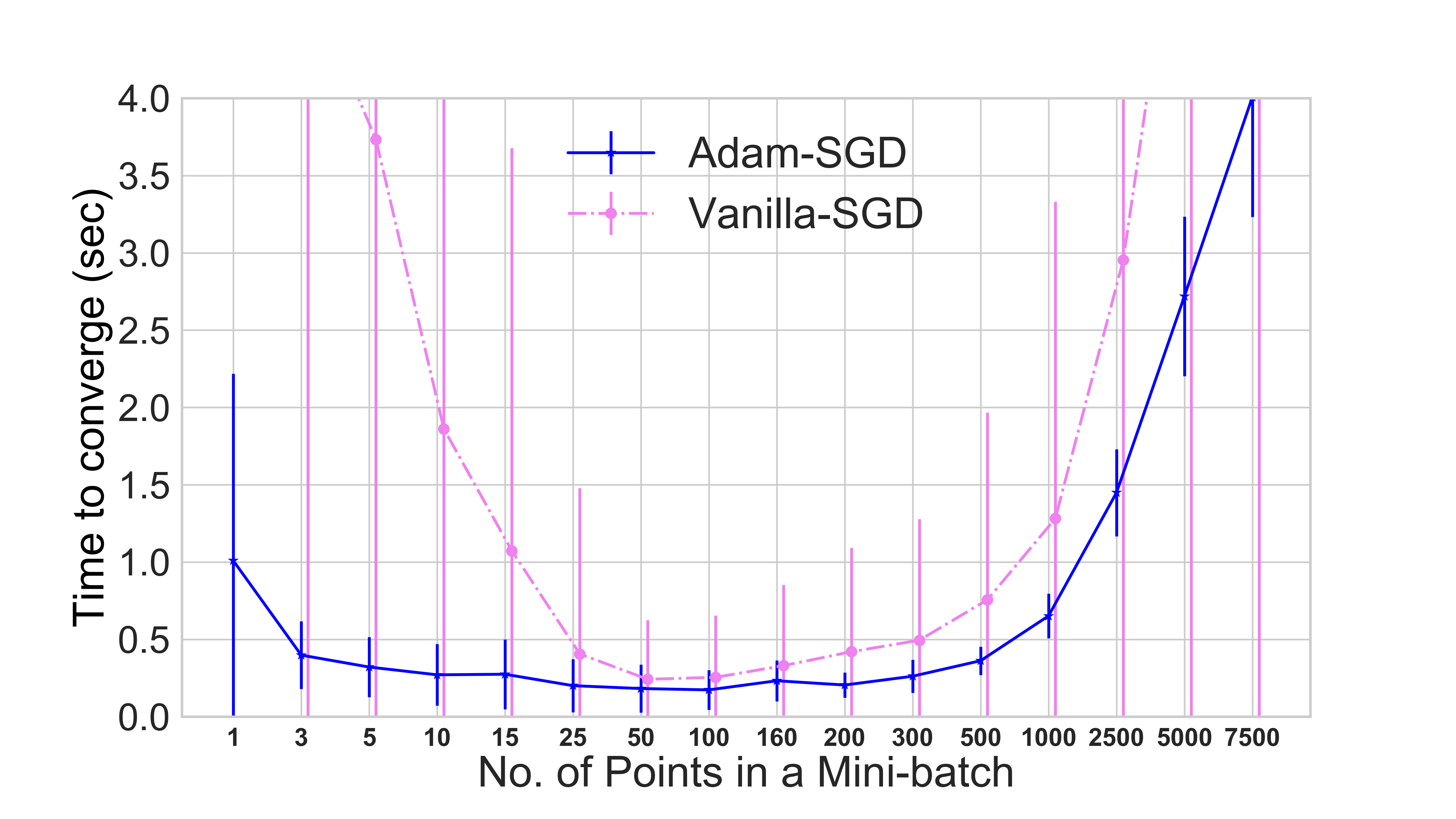}

\caption{Effect of batch-size on the performance of SGD-ICP algorithm in terms of accuracy and run-time using a fixed step-size (vanilla-SGD) and ADAM (Adam-SGD). These plots show that both vanilla-SGD and Adam-SGD work best with a batch-size of 50 to 300 points. Moreover, Adam-SGD is overall faster compared to vanilla-SGD.}
\label{fig:batch_size}
\end{figure*}

Figure \ref{fig:batch_size} shows the number of points in a mini-batch along the X-axis and a comparison of vanilla-SGD against Adam-SGD in terms of corresponding translational and rotational error in meters (m) and radians (rad) respectively, run-time in seconds (sec) and number of iterations along the Y-axis. This plot shows that a mini-batch containing 15 or more points gives the same solution quality. However, a bigger mini-batch containing 300 or more points needs more time to converge. This increases in the run-time can be attributed to the longer time consumed by data-association computation for larger mini-batches in each iteration. 

Comparison of Vanilla-SGD against Adam-SGD indicates that both perform equally well in terms of the solution quality, except that vanilla-SGD could not converge to the right solution for a mini-batch containing less than 15 points. Overall, Adam-SGD is superior to vanilla-SGD in terms of run-time efficiency as it converges in fewer iterations and in less time. The reason of superior performance of Adam-SGD is likely the smarter step-size which considers an exponentially decaying average of both past gradients and squared gradients in each dimension independently, resulting in quicker convergence.

\subsection{Velodyne Data}

So far all experiments are conducted on data gathered with a Kinect, which results in dense and evenly spaced point clouds with small ranges. In this experiment we use data gathered by a Velodyne from the KITTI dataset. This data has long range readings with highly variable point density. All methods use the exact same parameters they used in the previous experiments, with the goal being to see the impact of these parameters when moving between data from different sensors.

\begin{table}[bt]
	\centering
	
	\caption{Alignment errors on the KITTI dataset}
	\label{tab:kitti}
	
	\begin{tabular}{lrr}
	    \toprule
		Method & Translational & Rotational\\
		       & Error (m)   & Error (rad)\\
		\midrule
		SGD-ICP   & $1.2e^{-5} \pm 1.76e^{-5}$ & $2.4e^{-6} \pm 5.78e^{-6}$\\
		ICP       & $8.986     \pm 7.366$      & $0.152     \pm 0.116$ \\
		LIB-POINT & $0.003     \pm 0.001$      & $0.0004     \pm 0.00001$ \\
		LIB-PLANE & $0.108     \pm 0.589$      & $0.002     \pm 0.014$ \\
		GICP      & $9.442     \pm 6.911$      & $0.165     \pm 0.115$ \\
		\bottomrule 
	\end{tabular}
\end{table}

The results presented in Table \ref{tab:kitti} show that SGD-ICP has the lowest error, while standard ICP and GICP fail to produce usable results. A possible explanation for the poor performance of ICP and GICP is that they require proper parameter tuning to obtain good performance. On other other hand, SGD-ICP, LIB-POINT, and LIB-PLANE all perform well. The scaling of point clouds performed by SGD-ICP enables our method to handle Velodyne scans in the same manner as Kinect data. This scaling removes the need to perform the usually critical step-size tuning of SGD based methods. The random sub-sampling of points performed by the libpointmatcher methods renders them robust to the changes in the point clouds in addition to improving run-time.

In summary, these experiments demonstrate that SGD-ICP produces alignment errors comparable to standard ICP at a significantly reduced computational cost. In comparison to other methods that attempt to improve the run-time of ICP, the stochastic sampling of SGD-ICP provides the advantage that no a-priori data reduction, which can lead to loss of information, needs to be performed. 

However, SGD-ICP still shares the same drawbacks as standard ICP in that it is sensitive to outliers in case of partial overlap which can lead to wrong data associations, as discussed in the result of initial guess impact on the performance of the methods. This is an inherent characteristic of the point-to-point distance based data association.


\section{Conclusion}

In this paper we present an efficient and accurate ICP variant, called SGD-ICP, which employs stochastic gradient descent (SGD) to solve the optimisation problem at the core of ICP. Solving the optimisation using stochastic gradient updates results in significantly faster run-time without any loss in the quality of the final transformation estimate. Comparisons with other common methods show that our proposed method is faster, as accurate as standard ICP and other popular ICP variants, and easily applicable to point clouds from different sources without additional parameter tuning.
\addtolength{\textheight}{-14cm}
\section *{Acknowledgements}
This research is supported by an Australian Government Research Training Program (RTP) Scholarship.


\bibliographystyle{plain}
\bibliography{library4}

\end{document}